\documentclass{IEEEcsmag}

\usepackage[colorlinks,urlcolor=blue,linkcolor=blue,citecolor=blue]{hyperref}
\expandafter\def\expandafter\UrlBreaks\expandafter{\UrlBreaks\do\/\do\*\do\-\do\~\do\'\do\"\do\-}
\usepackage{upmath,color}
\usepackage{algorithm} 
\usepackage{algpseudocode} 
\usepackage{multirow}
\jvol{XX}
\jnum{XX}
\paper{8}
\jmonth{April}
\jmonth{}
\jname{Submitted to IEEE Micro}
\jtitle{Treasure What You Have}
\pubyear{2023}

\usepackage{tikz}

\renewcommand{\baselinestretch}{.985}

\begin{document}

%\sptitle{Article Type: Description  (see Introduction for more detail)}

\title{\textit{Treasure What You Have:} Exploiting Similarity in Deep Neural Networks for Efficient Video Processing \\
}

%\author{Hadjer Benmeziane\textsuperscript{1}, Halima Bouzidi\textsuperscript{1}, Hamza Ouarnoughi\textsuperscript{1, 2}, Ozcan Ozturk\textsuperscript{3} and Smail Niar\textsuperscript{1, 2}}

%\affil{\textsuperscript{1}Univ. Polytechnique Hauts-de-France, CNRS UMR8201 LAMIH,  Valenciennes, France}
%\affil\textsuperscript{2}{\textsuperscript{2} 
%INSA Hauts-de-France, Valenciennes, France}
%\affil\textsuperscript{3}{\textsuperscript{3} Computer Engineering Department, Bilkent University, Ankara, Turkey}

\DeclareRobustCommand*{\IEEEauthorrefmark}[1]{%
  \raisebox{0pt}[0pt][0pt]{\textsuperscript{\footnotesize #1}}%
}

\author{\IEEEauthorblockN{Hadjer Benmeziane\IEEEauthorrefmark{1}\textsuperscript{\textsection},
Halima Bouzidi\IEEEauthorrefmark{1}\textsuperscript{\textsection}, Hamza Ouarnoughi\IEEEauthorrefmark{1}\,\IEEEauthorrefmark{2}, 
Ozcan Ozturk\IEEEauthorrefmark{3} and Smail Niar\IEEEauthorrefmark{1}\,\IEEEauthorrefmark{2}} \\
\IEEEauthorrefmark{1}Univ. Polytechnique Hauts-de-France, CNRS UMR8201 LAMIH,  Valenciennes, France \\
\IEEEauthorrefmark{2}INSA Hauts-de-France, Valenciennes, France \\
\IEEEauthorrefmark{3}Computer Engineering Department, Bilkent University, Ankara, Turkey
}

\markboth{IEEE Micro : Special Issue on “tinyML”}{Special Issue on “TinyML”}

\begin{abstract}
%\looseness-
Deep learning has enabled various Internet of Things (IoT) applications. Still, designing models with high accuracy and computational efficiency remains a significant challenge, especially in real-time video processing applications. Such applications exhibit high inter- and intra-frame redundancy, allowing further improvement. This paper proposes a similarity-aware training methodology that exploits data redundancy in video frames for efficient processing. Our approach introduces a per-layer regularization that enhances computation reuse by increasing the similarity of weights during training. We validate our methodology on two critical real-time applications, lane detection and scene parsing. We observe an average compression ratio of approximately 50\% and a speedup of $\sim 1.5x$ for different models while maintaining the same accuracy.
\end{abstract}

\maketitle

\begingroup\renewcommand\thefootnote{\textsection}
\footnotetext{Denotes Equal Contribution}
\endgroup

%%% DL & HW
\chapteri{T}he advent of \textit{Deep Learning} (DL) has led to remarkable advancements in a wide range of applications, including natural language processing, computer vision, and speech recognition. 
In particular, DL has become an essential tool for enabling various Internet of Things (IoT) applications, where smart devices and sensors generate massive amounts of data that require processing and analysis in real-time. 
These applications include smart homes, autonomous vehicles, wearable devices, and smart cities, among others. 

%%% CNN
Following their success, DL models' design keeps evolving to dig for more and more performance. Along the way, Convolutional Neural Networks (CNN) have been predominant in computer vision tasks for a long time. 
The convolutional mechanisms and their variants (e.g., separable, depth-wise) can quickly learn the semantic features from data using the properties of locality and translation invariance. CNN typically operates on \textit{receptive fields} of the input feature maps using a set of filters. However, these receptive fields are not always activated and do not contribute equally to the feature extraction process, which also endures many redundant computations. 

%%% ViT
Recently, \textit{Vision Transformers} (ViT) have shown state-of-the-art results in computer vision tasks, including image classification and object detection. 
ViTs are primarily based on the multi-head self-attention operation (MHSA)~\cite{dosovitskiy2020image}. 
Nevertheless, the high effectiveness of ViTs comes at the cost of high computational complexity and memory footprint that burden their deployment on resource-constrained devices. Besides, increasing the number of heads in the attention operation promotes redundancy in computation.
%\textcolor{red}{needs to be reformulated ->} Besides, computation redundancy is encouraged in the attention operation by increasing the number of heads~\cite{}.

%%% NN optimization approaches
%Along this line, 

%Generally, 
\begin{figure*}
    \centering
    \includegraphics[width=0.9\textwidth]{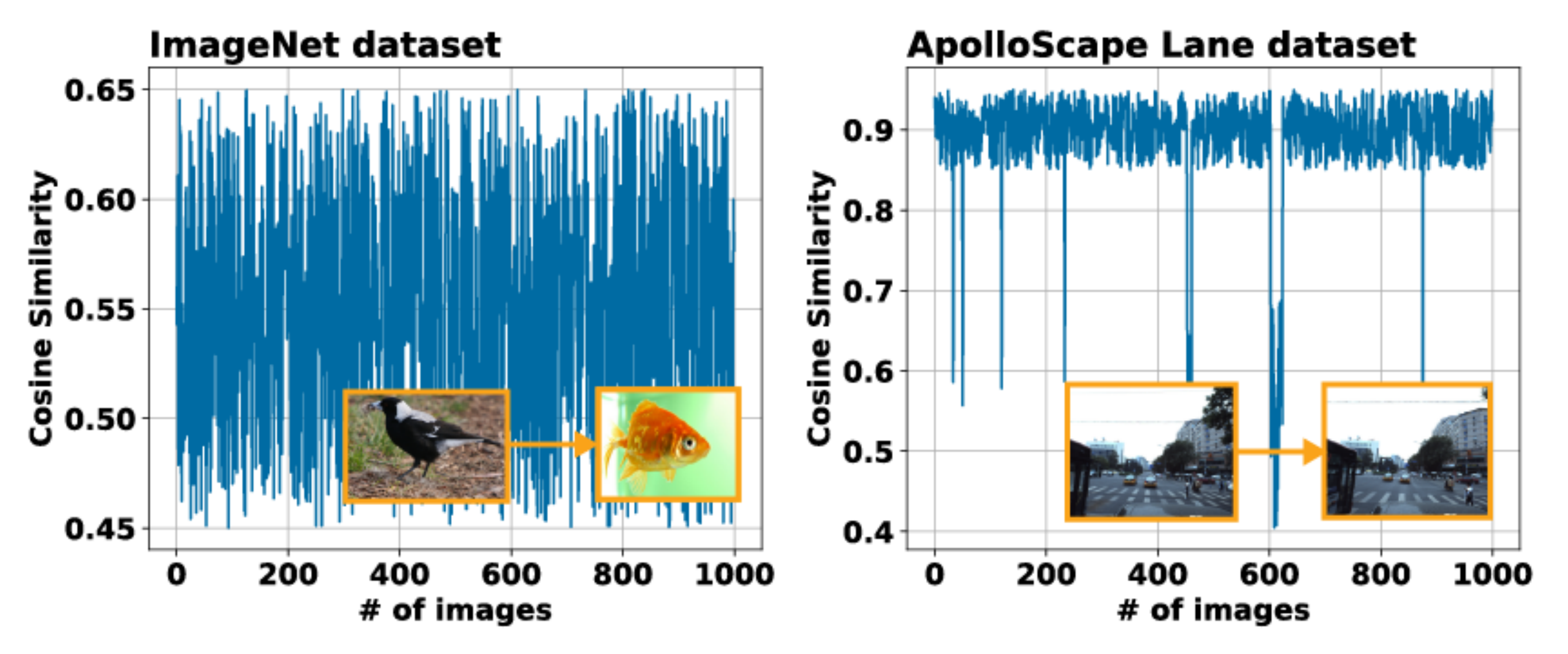}
    \caption{Inter-input cosine similarity in image-based datasets: ImageNet vs. a video-streaming dataset, lane detection in ApolloScape. 
    %\textcolor{red}{I suggest to make the figure on 2 columns. It is a good introductory figure, especially for Micro magazine and figure size doesn't matter.}
    }
    \label{fig:1}
\end{figure*}%\begin{enumerate}
 %   \item \textit{Quantization} that reduces the numerical precision of weights and feature maps to balance representational power and efficiency. %While \textit{FP16} and \textit{INT8} remain the most adopted quantization types, mixed-precision quantization has started gaining more attention lately \cite{wang2019haq}.
 %   \item 
 %\textit{Pruning} that removes the neurons with small saliency.
 %(i.e., contribution to the prediction task). 
%In the literature, we distinguish the unstructured and structured pruning schemes. The former removes neurons with small saliency no matter where they occur, whereas the latter removes a group of parameters (i.e., groups of filters). 

 %\item \textit{Knowledge distillation} that aims to transfer the knowledge from an overparameterized neural network (i.e., teacher) to another one (i.e., student) with fewer parameters and computational complexity. 
%The approach leverages the student's training on the soft probabilities generated by the teacher. 
%\end{enumerate}
Chasing the sparsity in DL model weights and feature maps has become a viable direction to reduce redundancy and use the computation budget wisely. 
Three main model compression techniques exist: quantization, pruning, and knowledge distillation. These techniques are usually adopted for DL models' deployment on tiny devices. % resource-constrained hardware devices (e.g., tiny micro-controllers). 
To further enhance the inference time, energy consumption, and memory footprint, researchers investigated the intra-inputs similarity. \textit{Computation reuse} is a novel optimization strategy for this purpose. 
It involves reusing the results of previous computations to avoid redundant computations, thus improving the model's efficiency. 
Computation reuse can be achieved through various methods, such as weight sharing, parameter tying, and feature map reuse. 
%DL models can reduce their memory requirements and computational load by reusing previously computed results. 
Data similarity typically manifests as repeated, irrelevant, or over-detailed information that may add extra processing overhead and cause the \textit{overthinking} phenomena that may mislead the inference process.
%Nonetheless, the high entropy that characterizes real-world input data makes it hard to straightforwardly predict and remove similarity at runtime in a way that pushes up the processing efficiency without drastically sacrificing accuracy. This is mainly due to the computational complexity required to detect such similarities. Therefore, researchers have turned to designing specialized accelerators that can quickly and accurately detect similarities in parallel and avoid redundant computations at runtime.

Video processing is an ideal application for computation reuse thanks to the spatiotemporal dependence between sequential frames. 
It is then possible to eliminate duplication by skipping identical frames. 
%as it depicts a high probability of redundancy within sequential frames. 
Therefore, it may gain an advantage using the computation similarity induced by DL models at either weight or data level. 
%Furthermore, because of the spatiotemporal dependence between the video frames, 
However, because similarity metrics may neglect finer-grained information, skipping at runtime is challenging, especially for critical real-time applications, such as autonomous driving systems. 
Thus, designing an efficient computation reuse method without decreasing the accuracy is essential to preserve reliability and robustness. 
Nonetheless, existing DL models are not designed nor trained with awareness of the similarity that may exist in runtime data. 
Thus, they cannot fully exploit the benefit of computation reuse if applied directly without further tuning.

\subsection{Motivational Example}\label{sec:motivation}
In Figure~\ref{fig:1}, we analyze the inter-input cosine similarity between 1000 images extracted from ImageNet and ApolloScape Lane detection~\cite{wang2019apolloscape} datasets, respectively. 
The correlation is much higher in the ApolloScape dataset due to the spatiotemporality dependence of video frames. 
On the other spectrum, the correlation in ImageNet does not follow any pattern, as images are entirely independent. 
This encourages us to apply computation reuse on video processing tasks. %Specifically, in this paper, we consider lane detection and scene parsing. NO NEED TO TALK ABOUT THE USE CASE. AS IT WILL BE DETAILED IN THE EXPERIMENTS.

\begin{figure*}
    \centering
    \includegraphics[ trim=4cm 0 0 0, clip,width=1.1\textwidth]{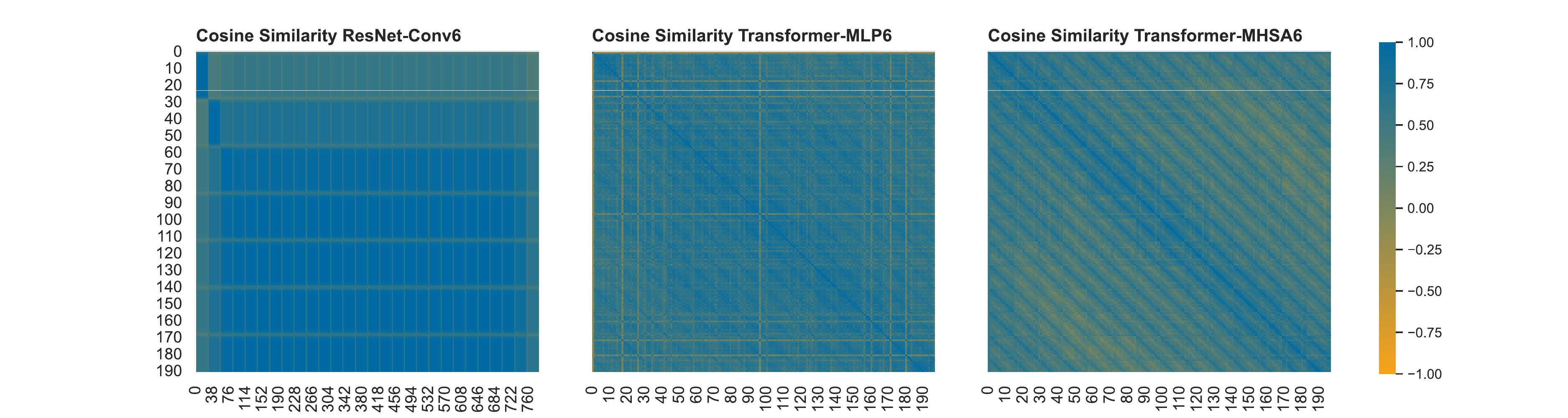}
    \caption{Intra-input cosine similarity in intermediate feature maps in Vision Transformer and ResNet. %\textcolor{red}{numbers on the x and y access and text are unreadable}
    }
    \label{fig:2}
\end{figure*}

Figure~\ref{fig:2}, on the other hand, analyzes the degree of computation similarity on different models. 
For this experiment, we take a transformer encoder from the base ViT~\cite{dosovitskiy2020image} and a convolutional layer from ResNet~\cite{he2016deep}. 
ViT encoder comprises an MHSA followed by a multi-layer perception (MLP). We generally observe a higher correlation in convolution than in the attention mechanism. 
This is mainly due to the spatial kernel shifting applied by the convolution. 
More so, this similarity ratio differs from one layer to another. Convolutional and MHSA layers exhibit lower correlation and similarity for deeper layers. 
This brings the question of how much the different CNN and ViT layers benefit from similarity reduction.

\textbf{Research questions:} Following the observations above, we formulate our research questions as follows:
\begin{itemize}
    \item How to exploit data similarity to boost computation efficiency while persevering DL models' effectiveness and representational power?
    \item How to scale DL model weights and architecture to exploit the full potential of both model and data redundancy?
\end{itemize}

\subsection{Proposed Contributions}
In this paper, we present the following contributions:
\begin{itemize}
    \item We propose a novel training approach that benefits from the intrinsic similarity of the training data to scale DL model weights and increases the computation reuse ratio during inference.
    \item We introduce an efficient and hardware-friendly inference method that detects data similarity on-the-fly using the Locality Sensitive Hashing (LSH).
    %which is hardware-friendly and enables computation reuse.
    \item To optimize the LSH hyperparameters concerning DL model's architectural features, we employ a Multi-objective Bayesian optimization algorithm that balances accuracy and compression ratio.
    \item We validate our proposed techniques on various DL models, including CNNs and ViTs, on various autonomous driving perception tasks from the \textit{ApolloScape} dataset.
\end{itemize}

\section{Related Works}
In literature, computation redundancy removal has been widely exploited for CNNs. \textit{SimCNN}~\cite{janfaza2021simcnn} has proposed a similarity-based convolution that detects data redundancy in filters and within receptive input data fields. These techniques allow, for fewer computations, improvements in training and inference for image classification. Other works, such as~\cite{buckler2018eva2}, targeted video streaming applications and decided to leverage the spatiotemporal correlation observed in video frames. Those works exploited the inter-frames data locality to reuse the prediction results or update the receptive fields when similarity is low. However, they depend on fixing a similarity threshold to decide whether to reuse previous results, making them less flexible and generalizable.
In contrast, \textit{Deep Reuse}~\cite{ning2019deep} proposes to apply data clustering with the LSH algorithm~\cite{andoni2015practical} on the fly to reduce the number of input activations and process only the centroids of the obtained clusters. Following this approach, no threshold must be set a priori as the clustering algorithm automatically detects data redundancy. The merit of this approach has been demonstrated for CNN by~\cite{ cicek2022energy}, showing the high energy efficiency gains.

\begin{table}[ht!]
\centering
\caption{Comparison between Related-works and ours. 
} 
\fontsize{9}{9}\selectfont
\scalebox{0.75}{
\label{tab:sota_table}
\begin{tabular}{ccccccc}
\hline
\textbf{Related Work} & \textbf{LSH} & \textbf{Conv/FC} & \textbf{ViT} & \textbf{\begin{tabular}[c]{@{}c@{}}Intra\\ frame\end{tabular}} & \textbf{\begin{tabular}[c]{@{}c@{}}Inter\\ frame\end{tabular}} & \textbf{\begin{tabular}[c]{@{}c@{}}HW\\ aware\end{tabular}} \\ \hline
SimCNN                &              & x                &              & x                                                              &                                                                & x                                                           \\ \hline
SumMerge              &              & x                &              & x                                                              &                                                                & x                                                           \\ \hline
Deep reuse            & x            & x                &              & x                                                              & x                                                              &                                                             \\ \hline
Cicek et al.          & x            & x                &              & x                                                              & x                                                              & x                                                           \\ \hline
Reformer              & x            &                  & x            & x                                                              &                                                                & x                                                           \\ \hline
\textbf{Ours}         & \textbf{x}   & \textbf{x}       & \textbf{x}   & \textbf{x}                                                     & \textbf{x}                                                     & \textbf{x}                                                  \\ \hline
\end{tabular}
}
\vspace{-1ex}
\end{table}

ViTs work on extracting visual patterns from images by performing pairwise dot-products within MHSA and MLP blocks. These dot products depict high computation redundancy, as not all tokens contribute equally to the output's prediction. Recent works~\cite{dong2022heatvit} leverage the underlying observation to reduce the number of computations by removing redundant heads. The closest line of research to ours is \textit{Reformer}~\cite{kitaev2020reformer} that introduces \textit{LSH attention}, an accelerated dot-product in attention layers using LSH clustering~\cite{andoni2015practical} to reduce the number of sequences being processed. However, our work differs from theirs as we investigate computation reuse opportunities in MHSA and MLP layers. More so, we propose optimizing the ViT model by adjusting weights regarding data similarity to leverage the full potential of both model and data similarities. Table~\ref{tab:sota_table} summarizes the critical differences between related works and ours.

\begin{figure}[t]
\centering
  \includegraphics[width=.5\textwidth]{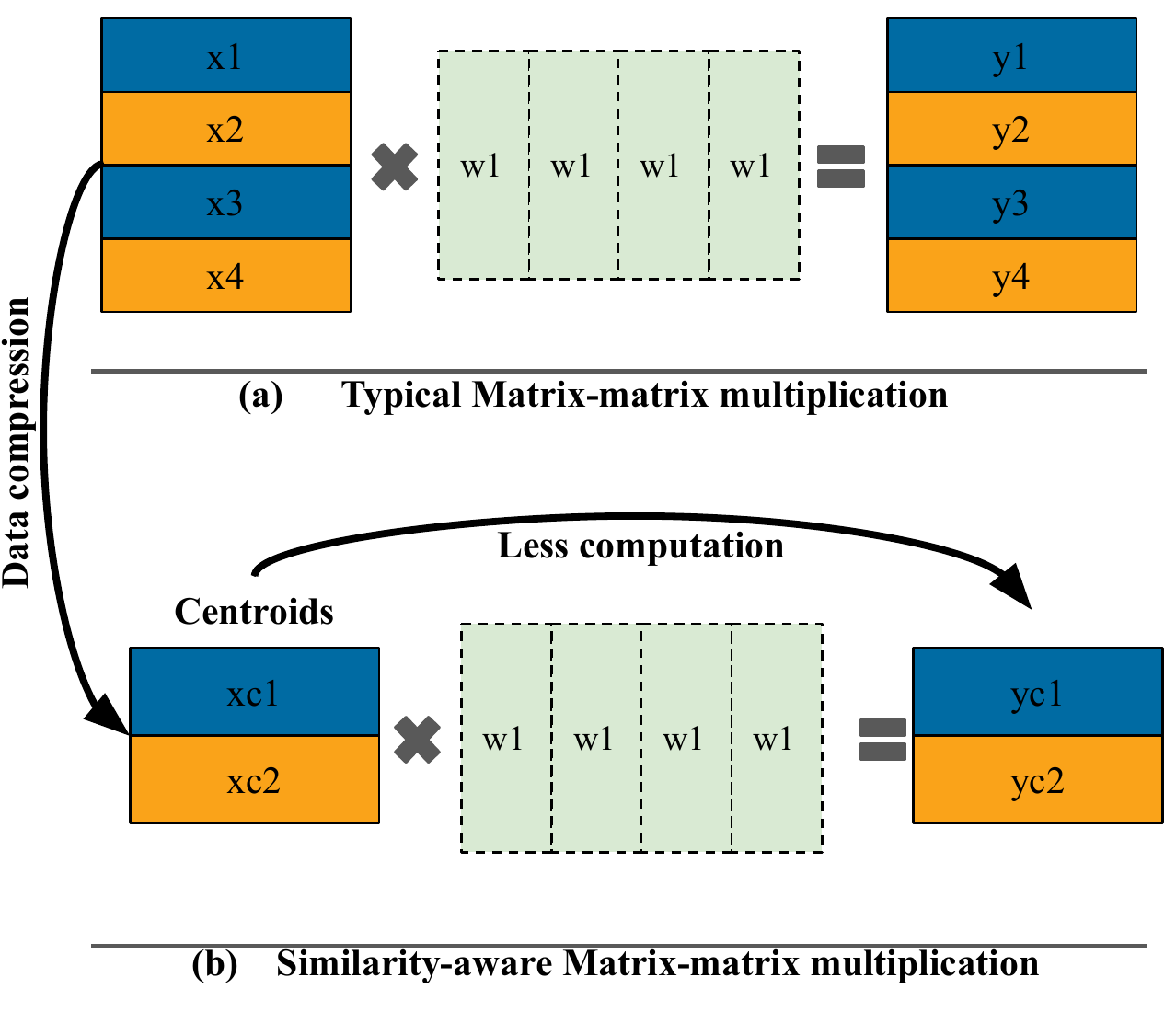}
  \caption{Illustration of computation reuse to reduce matrix-matrix multiplication complexity. Vectors of the same colors belong to the same cluster.}
  \label{fig:mm_multiply}
  \vspace{-1ex}
\end{figure}

\section{Preliminaries}
\subsection{Computation Reuse}
Matrix-matrix multiplication is crucial in various DL layers, like convolutional, fully connected, and attention layers. An image is usually transformed into a large input matrix $\mathcal{X}$ and then multiplied by a weight matrix $\mathcal{W}$ to produce an output matrix $\mathcal{Y}$, where each element $y_{i,j}$ of $\mathcal{Y}$ is the dot product of the $i$th row of $\mathcal{X}$ and $j$th column of $\mathcal{W}$. In video processing (Figure~\ref{fig:1}), this approach is inefficient as similar input rows $x_i$ are repeatedly processed with the same weights vectors $w_j$. To address this issue, we apply compression to the input matrix $\mathcal{X}$ using clustering algorithms like K-means or LSH, as shown in  Figure~\ref{fig:mm_multiply}. This step discovers sets of similar rows and generates {\it{centroid vectors}} to construct the new input matrix $\mathcal{X'}$. Then, we can use similarity-aware matrix multiplication to optimize matrix-matrix multiplication. Our work mainly focuses on similarity-aware optimization of matrix-matrix multiplication. Contrary to previous methods, we present a training methodology that enhances such similarities and exploits inter-frame computation redundancy. 
%\textcolor{red}{I suggest to add, if possible and not done elsewhere, a sentence to explain why conventional methods of MM compression are different from our work}

\subsection{Similarity Detection with LSH}~\label{sec:preliminaries}
\begin{figure}[t]
\centering
\includegraphics[width=0.5\textwidth]{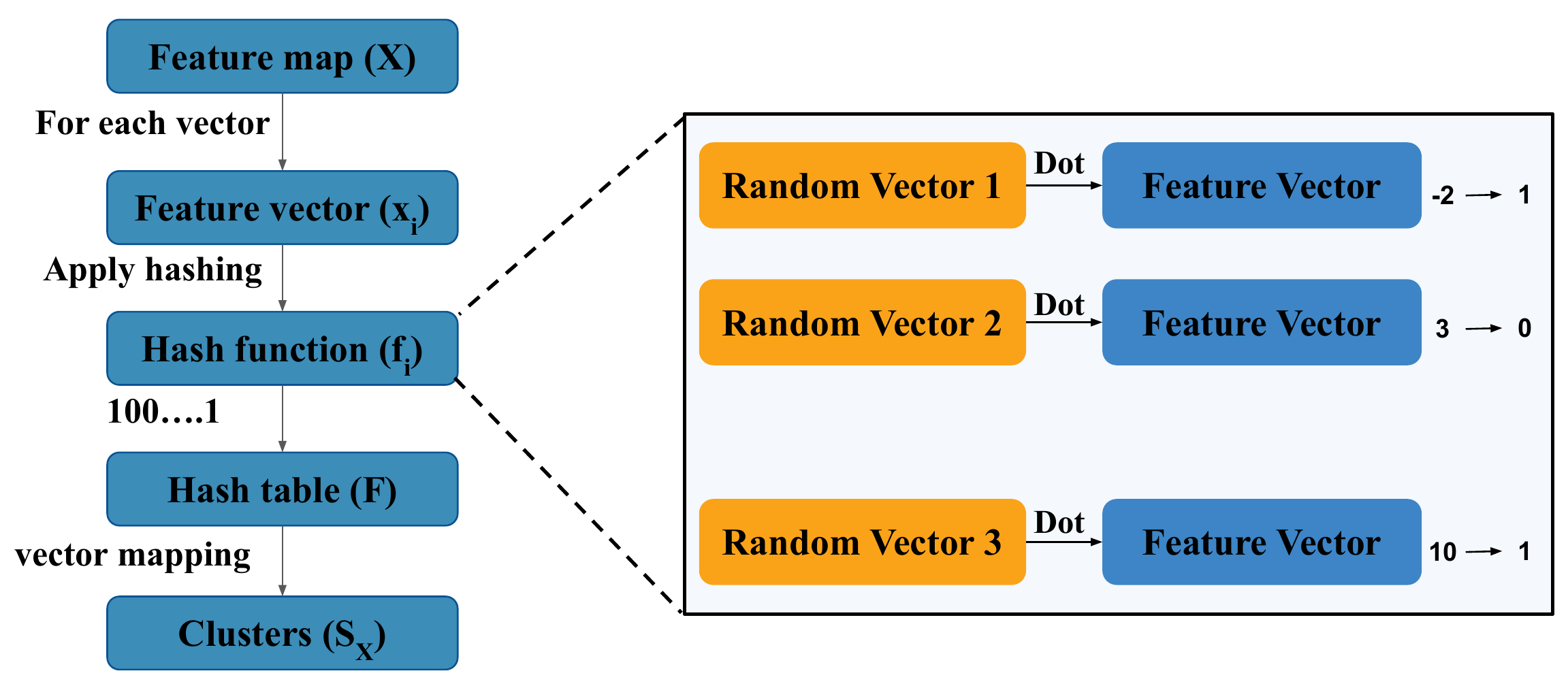} 
\caption{LSH algorithm workflow.}
\label{fig:lsh_worflow}
% \vspace{-1ex}
\end{figure}

LSH is a popular algorithm for quickly finding the nearest neighbors in high-dimensional feature spaces. The overall algorithm is presented in Figure~\ref{fig:lsh_worflow}. It is used to classify input vectors into clusters of similar neighbors based on  hash functions sensitive to locality. The hash function computes a dot product between feature vectors and $hash_{size}$ random projections. When multiple feature vectors have the same hash values, they are considered neighbors and assigned the same cluster $id$. To optimize LSH performance, two hyperparameters need to be optimized: $hash_{size}$ and $input_{dim}$. Increasing these parameters leads to more accurate clustering results but also higher computational complexity due to more dot products.

\begin{figure*}[t]
\centering
\includegraphics[width=.95\textwidth]{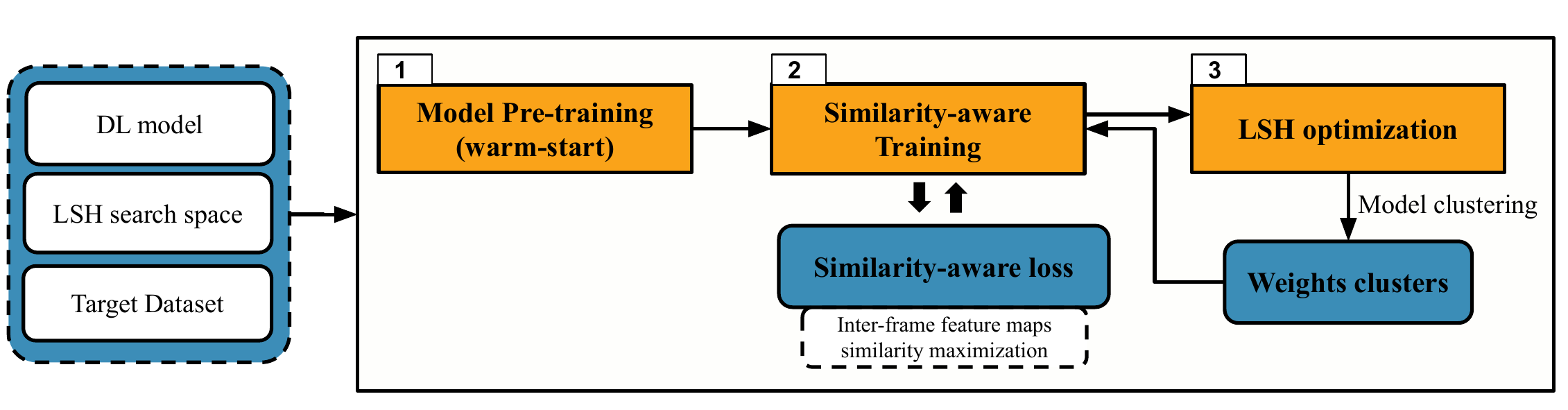} 
\caption{Overview of the proposed similarity-aware training approach.}
\label{fig:approach}
%  \vspace{-1ex}
\end{figure*}

\subsection{Computation Redundancy in CNN and ViT}
% Where is computation reuse applied in CNN and transformers
In DL models, redundancy manifests at model and data levels. Whereas various compression techniques have conquered model redundancy, data redundancy is still a significant optimization problem. We have synergically analyzed the redundancy spots within CNN and ViT layers. We observed that convolutional and transformer layers are the most rewarding as they are both computationally expensive, hold unnecessary weights, and process redundant data. 
A convolutional layer involves applying $K$ \textit{sliding window} filters of dimension $R \times S \times C$ to a $W \times H \times C$ dimension input feature map ($\mathcal{X}$), generating a $(W - R +1)  \times (H -S +1) \times K$ output ($\mathcal{Y}$). From a matrix-matrix multiply perspective, for each spatial dimension, this can be seen as the dot product between a vector of filters and a vector of input features as follows: 
\begin{align*}
\mathcal{Y}[(k,i,j)] = \sum_{c=0}^{C-1}\sum_{r=0}^{R-1}\sum_{s=0}^{S-1}F[(k,c,r,s)] * \mathbf{\mathcal{X}}[(c,i+r,j+s)] \\
\text{for} \quad 0 \le k < K, \quad 0 \le i \le W-R, \quad 0 \le j \le H-S
\label{eqn:conv}
\end{align*}
 Similarity can be exploited within the input feature vectors by removing the redundant ones and applying the LSH clustering on the feature vectors of $\mathcal{X}$ and operating on the centroids vectors only ($\mathcal{X'}$). 

Similarly, a ViT  
%ViT is characterized by transformer blocks composed of MHSA layers followed by an MLP. In particular, 
transformer block is computed as follows:
\begin{gather*} 
\texttt{Attn}(\mathbf{Q}, \mathbf{K}, \mathbf{V}) = \texttt{softmax}(\frac{\mathbf{QK}^T}{\sqrt{d_k}})\mathbf{V} \\
\text{head}_{i} = \texttt{Attn}(\mathcal{X}W_{i}^Q, \ \mathcal{X}W_{i}^K, \ \mathcal{X}W_{i}^V) \\
\texttt{MHSA}(\mathcal{X}) = \texttt{Concat}(\text{head}_{1}, ..., \text{head}_{H})W^O \\
Z = \texttt{MHSA}(\mathcal{X}) + \mathcal{X}, \quad \mathcal{Y} = \texttt{MLP}(\mathbf{Z}) + \mathbf{Z},
\label{eqn:attn}
\end{gather*} 
where Q, K, V are query, keys, and values matrices extracted from the same input feature map ($\mathcal{X}$) by applying linear projection using learnable weights ${W}_{i}^K, {W}_{i}^Q, {W}_{i}^V$ for each $head_i$. As these linear projections may cause a data distribution shifting and thus alter the data similarity within the original input feature map, we apply the LSH clustering in a more fine-grained granularity. We use the Q, K, and V matrices for each MHSA head. 

According to our preliminary analysis, compressing the Q, K, and V matrices allows for better computation reuse opportunities. In addition, a slight decrease in accuracy is obtained compared to operating on the original input feature map ($\mathcal{X}$), where the compression is more aggressive with a significant accuracy decrease. More so, we apply the LSH on the output of the MHSA (i.e., the input of MLP), noted as $Z$. MLP layers are computationally more expensive than MHSA and prone to data redundancy. In contrast to the approach proposed in \cite{kitaev2020reformer}, where the LSH is only applied on MHSA, our approach considers MLP in the loop, increasing the computation reuse ratio.
%\vfill{5cm}
\section{Proposed Approach}
\subsection{Functional Architecture of the Approach}
The overall process of our similarity-aware training is depicted in Figure \ref{fig:approach} and can be described as follows:
\begin{enumerate}
\item \textbf{Pre-training}: The first step is to pre-train a model on a large dataset. This enables a warm start to similarity-aware training. 

\item \textbf{Similarity-aware training} (SA-training): The next step is to fine-tune the pre-trained model on the target dataset. During this stage, we use a novel regularizer that minimizes the Kullback–Leibler (KL) divergence between the output feature maps of consecutive inputs. The training procedure is detailed below. 

\item \textbf{LSH optimization}: While training, we apply LSH to cluster the model's feature maps. The clustering is automatically applied just before the convolution layer, the MHSA, or the fully connected (FC) layer. 
%LSH is a technique used for finding similar items in high-dimensional spaces. In this case, it is used to cluster the model weights. 
However, LSH is computationally intensive and requires setting multiple hyperparameters per layer. We thus provide an optimized LSH hyperparameter tuning and acceleration. 

\item \textbf{Similarity-aware loss}: After LSH clustering, we compute the loss for each cluster. This loss is based on the difference between the weights in each cluster and the average weight for that cluster. The goal is to encourage the model to converge to a similar set of weights for each cluster.

\end{enumerate}
Finally, this process is repeated iteratively during training. The regularizer and LSH clustering are applied at each iteration to enhance the inherent similarity in the pre-trained model.

\subsection{Similarity-aware Training}\label{sec:training}
During the training process of a neural network, the set of trainable parameters, denoted by $\theta$, are adjusted to optimize the task-specific loss function.
%, such as cross-entropy for classification or mean squared error (MSE) for lane detection. %For lane detection, the MSE loss is commonly used to measure the difference between the predicted lane boundaries and the ground truth lane boundaries, which is computed by squaring the element-wise differences and averaging over all pixels in the image.
Our regularization term promotes computation reuse in deep learning models by encouraging similar output feature maps at each layer. When discussing similarity in this paper, we are referring specifically to the similarity between the rows constructing the clusters. In this context, similarity refers to the degree to which rows of inter-frames and inter-feature maps are clustered within the same cluster. %This allows the model to reuse computations and reduce the overall computational cost during both training and inference.

The regularization term is formulated using the KL-divergence, which measures the difference between two probability distributions. 
In our case, the first distribution is the output feature map of the $i$th layer, denoted by $FM_i$. The second distribution is the mean of the feature maps across the training dataset, represented by $\mu_i$. To compute the regularization term, we first flatten the output feature maps of each layer into a 1-dimensional vector, denoted by $f_i$ in Equation~\ref{eq:regularization}. 
%Then, the KL-divergence between the distribution of $f_i$ and the reference distribution $\mu_i$ is computed as:
%\begin{equation}
%D_{KL}(f_i || \mu_i) = \sum_j f_i(j) \log \frac{f_i(j)}{\mu_i(j)} + \sum_j \mu_i(j) \log \frac{\mu_i(j)}{f_i(j)},
%\label{eq:kl_div}
%\end{equation}
%where $j$ indexes the elements of the vectors $f_i$ and $\mu_i$. 

The regularization term $r$ is then defined as the sum of the KL-divergence for each layer $i$:
\begin{equation}
r = \sum_{i=1}^{L} D_{KL}(f_i || \mu_i),
\label{eq:regularization}
\end{equation}
where $L$ is the total number of layers. During training, the regularization term is added to the task-specific loss function
%, such as the MSE loss for lane detection, 
to form the total loss function to optimize.

In addition to enhancing the similarity between feature maps within a single frame, our regularization method can also be extended to improve the similarity between feature maps across different frames. %In real-time video applications, maintaining consistency and coherence between the outputs of consecutive frames is often desirable. 
%By encouraging the model to learn more similar feature maps between frames, o
%Our regularization term can consequently improve the temporal coherence of the model's output, leading to smoother and more stable results over time. 
%This can be achieved by adding a regularization term to the loss function that encourages similarity between feature maps across consecutive frames. 
%Specifically, we penalize the difference between the feature maps of two successive frames, such as by using a mean squared error loss or another suitable metric. 
Equation~\ref{eq:final} shows the regularization term that enhances feature maps' similarity between multiple frames. 

\begin{figure}[ht]
\centering
\includegraphics[width=.5\textwidth]{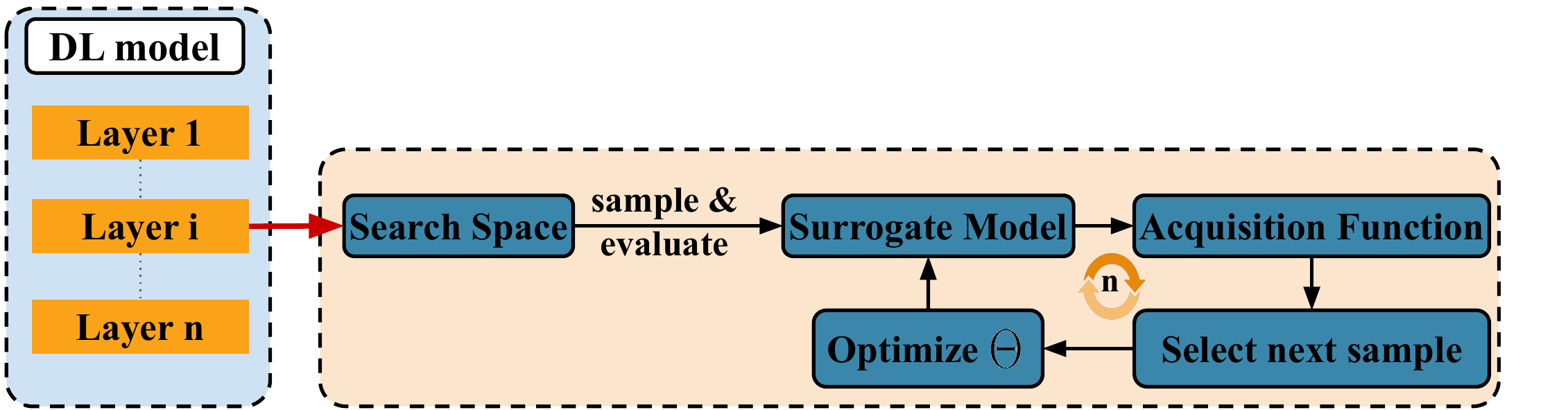} 
\caption{Parallel Multi-Objective Bayesian Optimization (MOBO) approach to optimize LSH hyperparameters.}
\label{fig:lsh_opt}
  \vspace{-1ex}
\end{figure}

\begin{equation}
r_t = \sum_{i=1}^{L-1} \sum_{j=i+1}^{L} \lambda_{ij} \cdot D_{KL} (FM_{i,t}||FM_{j,t+1})
\label{eq:final}
\end{equation}

% Please add the following required packages to your document preamble:
% \usepackage{multirow}
\begin{table*}[t]
\centering
\caption{Overall results on Lane Detection and Scene Parsing. The inference speedup is computed on Raspberry Pi3.}
\label{tab:results}
\begin{tabular}{ll|p{1cm}|p{1cm}|p{1cm}|p{1cm}|p{1cm}}
\hline\hline
\multicolumn{1}{l|}{\textbf{Dataset/Task}}                    & \textbf{Model}           & \multicolumn{1}{p{1.5cm}|}{\textbf{Original F1-score}} & \multicolumn{1}{p{1.5cm}|}{\textbf{SA-trained F1-score}} & \multicolumn{1}{p{2cm}|}{\textbf{Inference Speedup}} & \multicolumn{1}{p{2cm}|}{\textbf{Compression Ratio}} & \textbf{Training time} \\ \hline\hline
\multicolumn{1}{l|}{\multirow{6}{*}{Lane Detection}} & Resnet18        & 62,1                                   & 68,54                                    & 1,59                            & 43,10                           & 2h34          \\ \cline{2-7} 
\multicolumn{1}{l|}{}                                & Resnet18-8bits  & 58,3                                   & 65,32                                    & 1,01                           & 45,2                                   & 1h46          \\ \cline{2-7} 
\multicolumn{1}{l|}{}                                & ResNeSt         & 78,54                                  & 77,45                                    & 2,66                          & 61,33                           & 2h51          \\ \cline{2-7} 
\multicolumn{1}{l|}{}                                & ResNeSt-8bits   & 78,42                                  & 77,48                                    & 2,31                                   & 72,5                                   & 2h12          \\ \cline{2-7} 
\multicolumn{1}{l|}{}                                & Swin            & 75,62                                  & 76,31                                    & 1,68                                   & 62,3                                   & 2h32          \\ \cline{2-7} 
\multicolumn{1}{l|}{}                                & Swin-8bits      & 75,6                                   & 75,5                                     & 1,45                                   & 65,21                                  & 2h03          \\ \hline\hline \\ \hline\hline
                                             \multicolumn{1}{l|}{\textbf{Dataset/Task}}           &        \textbf{Model}          & \multicolumn{1}{p{1.5cm}|}{\textbf{Original mIoU}}    & \multicolumn{1}{|p{1.5cm}|}{\textbf{SA-trained mIoU}}     & \multicolumn{1}{p{2cm}|}{\textbf{Inference Speedup}} & \multicolumn{1}{p{2cm}|}{\textbf{Compression Ratio}} & \textbf{Training time} \\ \hline \hline
\multicolumn{1}{l|}{\multirow{4}{*}{Scene Parsing}}  & Resnet-38       & 63,2                                   & 65,12                                    & 2,4                                    & 51,2                                   & 1h21          \\ \cline{2-7} 
\multicolumn{1}{l|}{}                                & Resnet-38-8bits & 63,5                                   & 64,52                                    & 1,51                                   & 56,34                                  & 1h15          \\ \cline{2-7} 
\multicolumn{1}{l|}{}                                & PSANet          & 82,51                                  & 81,32                                    & 2,1                                    & 23,65                                  & 2h10          \\ \cline{2-7} 
\multicolumn{1}{l|}{}                                & PSANet-8bits    & 82,5                                   & 82,51                                    & 1,36                                   & 34,5                                   & 1h54          \\ \hline\hline
\end{tabular}
\end{table*}

Here, $FM_{i,t}$ and $FM_{j,t+1}$ represent the output feature maps of layers $i$ and $j$ at time $t$ and $t+1$, respectively. The regularization parameter $\lambda_{ij}$ controls the regularization strength between layers $i$ and $j$. By penalizing the difference between the feature maps of consecutive frames, this regularization term encourages the model to learn more similar features over time, leading to smoother and more coherent output. %Note that this modification is just one example of how the regularization term could be extended to incorporate temporal consistency, and other modifications are also possible depending on the specific application and model architecture.

During training, we apply similarity detection and computation reuse to measure the cross-entropy loss using centroids multiplication. Algorithm~\ref{alg:training} details the training algorithm. 

\begin{algorithm}
    \caption{Similarity-Aware Training Algorithm}
    \label{alg:training}
    \begin{algorithmic}
        \State \textbf{Inputs: } Training dataset $D$, NN architecture $f(x, theta)$, regularization strength $\lambda$, time regularization strength $\lambda\_{t}$, learning rate $\alpha$, number of training iterations $epochs$

        \State Pre-training the weights $\theta$
        \State Preprocess the training dataset $D$ to extract input feature maps for each sample
        \For{epoch = 0 to epochs:}
            \State Apply LSH and store rows and cluster ids.
            \For{sample $x$, $label$ in $D$:} 
                \State $output \gets f(x)$ using LSH clusters
                \State $L\_task \gets Loss(output, label)$
                \State R = 0
                \For{each layer}
                    \State Compute $r$ using Equation 1
                    \State Compute $r_t$ using Equation 2
                    \State $R += \lambda*r + \lambda_{t}*r_t$
                \EndFor
                \State update $\theta$
            \EndFor
        \EndFor
    \end{algorithmic}
\end{algorithm} 

\subsection{LSH Hyper-parameters Tuning}\label{sec:tuning}
As described in the \textit{preliminaries} section, two hyperparameters must be tuned to apply LSH: the number of hash functions ($hash_{size}$) and the size of feature vectors ($input_{dim}$). In the motivational experiment section, we noticed that different model layers depict different cosine similarities. This indicates that each layer will require different LSH hyperparameters.

The hyperparameter tuning methodology employed in this study utilizes a nested Multi-Objective Bayesian Optimization (MOBO) to fine-tune the LSH hyperparameters of each layer in the model. Figure~\ref{fig:lsh_opt} illustrates this optimization process. 
%MOBO is a probabilistic approach that models the objective function. 
We ran the MOBO in parallel for every layer to speed up the optimization process. 
%in which clustering was applied. 
Because the optimization is done layer-wise, the validation accuracy cannot be used as an objective. Instead, we use the MSE between the original layer's output and the similarity-aware layer's output. We also add the compression ratio as a second objective to maximize. The compression ratio is computed by dividing the original input and centroid sizes. The overall objective function is denoted as $\Theta$ and formulated in Equation~\ref{eq:4}, where $\sigma$ corresponds to the compression ratio, 
%$MSE$ denotes the normalized mean-squared error, 
and $a$, $b$ denotes the input after and before compression, respectively. 
\begin{equation}
\label{eq:4}
    \Theta = \frac{MSE(a, b)}{\sigma(a, b)}
\end{equation}
The optimization process proceeds as follows:
\begin{enumerate}
    \item Choose an initial set of parameter values to evaluate the objective function.
    \item Fit a Gaussian probabilistic model~\cite{gaussian} to the evaluated points. This model estimates the objective function of unexplored points in the parameter space.
    \item Use an acquisition function to select the next point to evaluate. The acquisition function balances the trade-off between exploration and exploitation. We use the expected improvement~\cite{eif}. 
    %for this purpose.
    \item Evaluate the objective function at the selected point.
    %and add it to the set of evaluated points.
    \item Update the probabilistic model with the new point and repeat steps 3-5 until a maximum number of evaluations, denoted as $n$, is reached.
\end{enumerate}

% Please add the following required packages to your document preamble:
% \usepackage{multirow}

\section{Evaluation}
We evaluate our proposed training methodology on the ApolloScape dataset for autonomous driving. We target \textit{lane detection} and \textit{scene parsing} as application cases. We compare the performance of our method against baseline models trained with and without our regularization. We also compare state-of-the-art methods for lane detection and scene parsing. We deployed the obtained model on Raspberry Pi3, a widely available and cost-effective option for running tinyML models. It also offers a low-power and compact computing platform suitable for running real-time applications. 

\subsection{Experimental Setup}
\paragraph{Dataset} 
Using ApolloScape 
%dataset is a large-scale, high-resolution dataset for autonomous driving research. It contains various challenging scenarios for autonomous vehicles. 
we have evaluated our approach on two sub-datasets for:  
%The dataset is created by Baidu's Apollo project.
%The goal of lane detection is to detect and track lane markings on the road. Lane detection typically involves processing images captured by a front-facing camera and generating pixel-level annotations for the lane markings in the image. \textit{The ApolloScape Lane Detection} dataset is a sub-dataset of ApolloScape, consisting of high-resolution images of urban road scenes captured by a front-facing camera. 
1) \textit{Lane detection} contains over 10,000 labeled images with pixel-level annotations for lane markings. 
%The dataset is split into a training set, a validation set, and a test set.
%Scene parsing, or semantic segmentation, is another essential task in autonomous driving systems. The goal of scene parsing is to assign a semantic label to each pixel in an image, indicating the object category it belongs to, such as road, building, pedestrian, or vehicle. \textit{The Scene Parsing} 
2) \textit{Scene parsing} comprises 38,000 high-resolution labeled images with pixel-level annotations for different object categories such as road, building, vegetation, and vehicles. 
%The dataset contains over 38,000 labeled images and is split into training, validation, and test sets.

\paragraph{Setup} 
We implement our method using the PyTorch framework and train all models on an NVIDIA GeForce RTX 3090 GPU. All models are initially pretrained with standard training methodology. If quantization is mentioned, the models are trained with quantization-aware training. For training, we use SGD with momentum as an optimizer and a learning rate of 0.01. We apply our regularization with a strength parameter $\lambda$ of 0.001 and use LSH to group similar feature maps in each layer after tuning its hyperparameters.
%using the methodology described in the Proposed Approach section. 
All models are trained for 50 epochs with a batch size of 16.

\subsection{Results}
Table~\ref{tab:results} summarizes the performance of different models on the ApolloScape validation set for Lane Detection and Scene Parsing. We compare the original pre-trained model against the same model with additional fine-tuning using SA-training. We also compare the quantized version of such a model to an SA-trained quantized version for each model. We apply an 8-bit quantization to both models' activations and weights. Quantization-aware training is applied before SA-training.

%\begin{figure}[t]
%    \centering
%    \includegraphics[width=0.5\textwidth]{figures/training_analysis.pdf}
%    \caption{Evolution of negative KL-divergence during training for each layer for five runs on Resnet18.}
%    \label{fig:kl}
%\end{figure}

For Lane Detection, our proposed SA-training methodology outperforms baseline models trained without LSH, achieving an average speedup of 1.78 and reducing the models' computations by 58\%. For the Resnet-18 and Swin models, the additional training induced an F1-score increase of 6\% and 1\%, respectively. In ResNeSt, however, the computation reuse impacted a 1\% accuracy drop. In the case of ResNeSt, its nested multi-scale feature design makes it more challenging to identify similar computations between different layers. As a result, some computations that are not genuinely similar might be grouped and shared during similarity-aware training, leading to a slight loss in F1-score. The quantization allows the computation reuse to achieve a higher compression ratio induced by the restricted range of weights and activation values. 

Similar conclusions can be deducted from the Scene Parsing results. SA-training achieved an average speedup of 1.8 and a compression ratio of 41\%. Except for PSANet, the additional training increased the final mIoU score. 

%Although applying the similarity check and LSH clustering during training can lead to a longer training time, we consider this methodology worthwhile due to the performance improvement achieved and the fact that it only requires a one-time training run.

We further analyze our regularization method's impact on the models' compression rate. We find that our approach achieves an average compression rate of 50\% across different models while maintaining the same level of accuracy as the baselines. This demonstrates the effectiveness of our method in enhancing computation reuse in deep learning models.

\begin{figure}[t]
    \centering
    \includegraphics[width=0.5\textwidth]{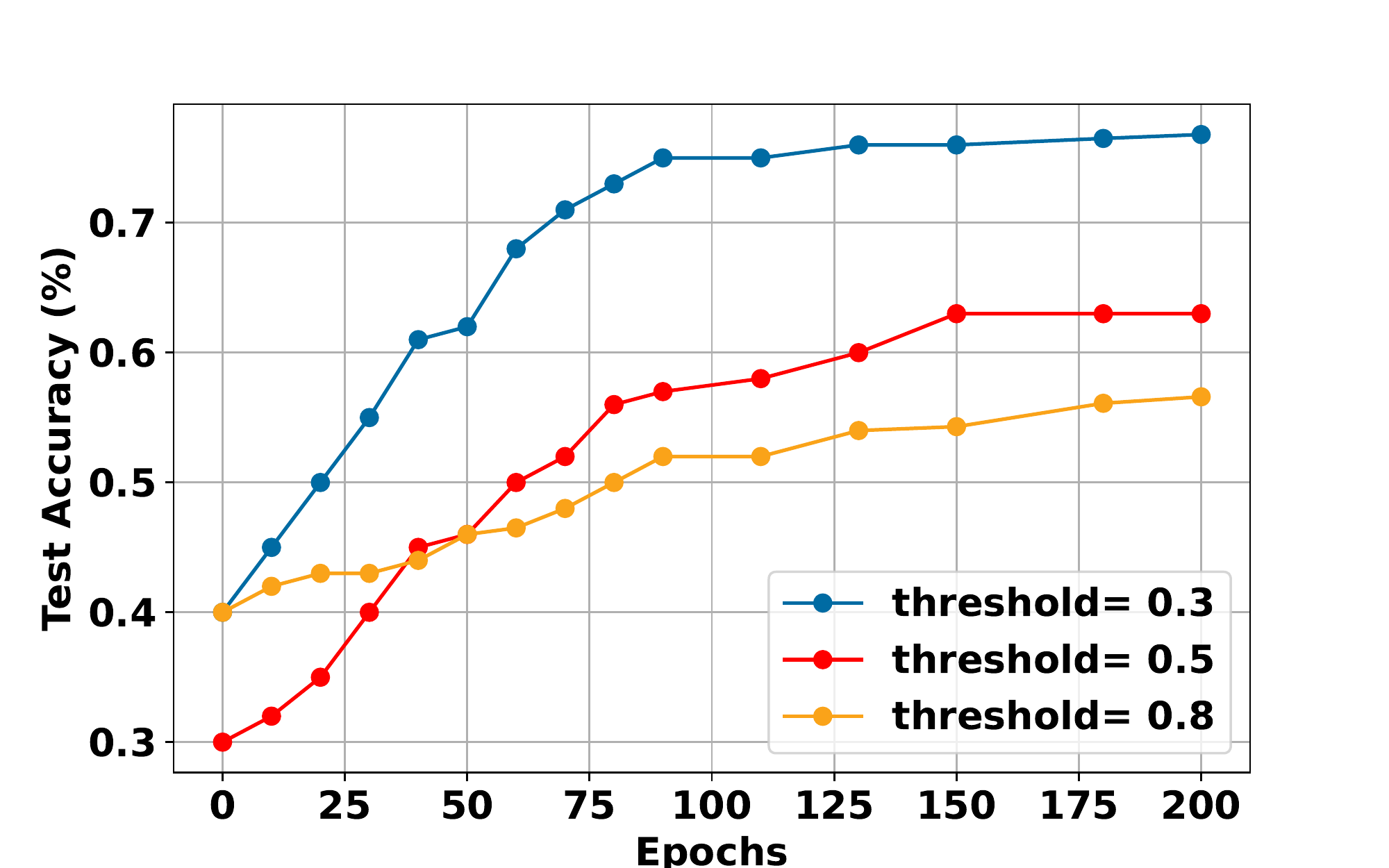}
     \caption{Training performance with different frame regularization strengths.}
     \label{fig:threshold}
\end{figure}

%Figure~\ref{fig:kl} shows the trend of similarity increase over the training epochs for each layer of the deep learning models trained using our proposed methodology. As can be seen from the figure, the similarity between feature maps of different layers gradually increases as the training progresses. This trend is observed for both ResNet and Swin architectures on the ApolloScape dataset for lane detection. The x-axis represents the number of training epochs, while the y-axis represents the similarity between feature maps. The increasing trend of similarity between feature maps indicates that our regularization approach combined with the LSH technique effectively enhances computation reuse in deep learning models, leading to better performance in real-time video applications such as lane detection.

We set a threshold on the number of frames used. To analyze the impact of different frame thresholds, we plotted the accuracy trend in Figure~\ref{fig:threshold}. The x-axis of the figure represents the number of training epochs, while the y-axis shows the model's accuracy. The line graphs in different colors indicate the accuracy trend for different threshold values. The results indicate that the model's accuracy increases as the threshold values increase, suggesting that more frames contribute to the training process, leading to better accuracy. The increasing trend of accuracy for higher threshold values demonstrates the effectiveness of our proposed methodology in enhancing the training process and improving the model's accuracy.

Finally, an ablation study revealed that training solely on inter-feature map similarity achieved a 20\% compression ratio with 1.2x speedup. However, training solely on inter-frame similarity resulted in a 33\% compression ratio with 1.7x speedup. Combining both approaches yielded better compression and faster inference.

\section{Conclusion}
This paper presents a novel similarity-aware training approach for Deep Neural Networks exploiting intra- and inter-frame similarity in video processing tasks to accelerate real-time computations. Specifically, we introduce a regularization approach that enhances weights similarity to align with input data similarity induced by the spatiotemporal correlation in video frames. 
We have shown that our approach can achieve up to $\sim$ \textbf{50\%} compression ratio on different models while enjoying desired accuracy levels on two autonomous driving tasks: lane detection and scene parsing.  

\bibliographystyle{IEEEtran}
\bibliography{references}

\noindent
\noindent
\textbf{Hadjer Benmeziane} received her Master's and Engineering degrees in Computer Science at the Higher National School of Computer Science, Algiers, Algeria, in 2020. She is currently pursuing a Ph.D. degree in Hardware-aware Neural Architecture Search at the  University Polytechnic Hauts-de-France, LAMIH/CNRS, Valenciennes, France.

\noindent
\textbf{Halima Bouzidi} 
received her Master's and Engineering degrees in Computer Science at the Higher National School of Computer Science of Algiers in 2020. She is currently pursuing a Ph.D. degree at the University Polytechnic Hauts-de-France of Valenciennes. Her research area encompasses ML design automation and energy-efficient edge AI.

\noindent
\textbf{Hamza Ouarnoughi} 
has been an associate professor in the Computer Science Department at INSA Hauts-de-France, and a researcher at LAMIH since 2018. He received his Ph.D. degree in computer science from The University of Western Brittany in 2017. His research interests are Cloud and Edge Computing, embedded systems, and AI.

\noindent
\textbf{Ozcan Ozturk} 
 has been on the faculty at Bilkent since 2008, where he is a Professor in the Department of Computer Engineering. His research interests include manycore accelerators, on-chip multiprocessing, computer architecture, and compiler optimizations. Dr. Ozturk received his Ph.D. in computer science and engineering from the Pennsylvania State University, an M.S. in computer engineering from the University Of Florida, and a B.Sc. degree in computer engineering from Bogazici University, Turkey.

\noindent
\textbf{Smail Niar} 
 received his Ph.D. in Computer Engineering from the University of Lille in 1990. Since then, he has been a professor at the University Polytechnique Hauts-de-France (UPHF) at LAMIH, a joint research unit between CNRS and UPHF.
 His research interests are AI/ML-based embedded systems, autonomous driving, HPC, and edge computing.

\end{document}